# Quadrature Oscillation System for Coordinated Motion in Crawling Origami Robot

Sean Liu[1], Ankur Mehta[2], Wenzhong Yan[3,†]

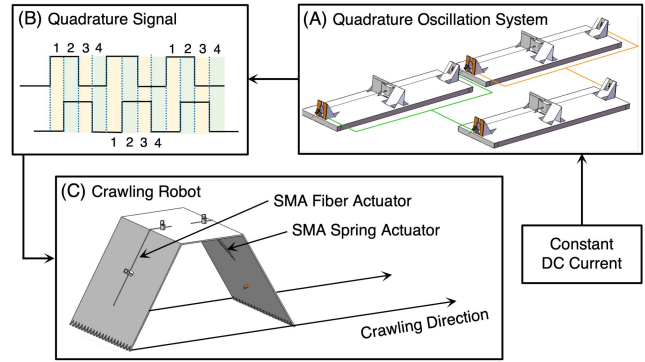

Figure 1. Schematic of a crawling robot driven by a quadrature oscillator with a constant DC power input. (A) Quadrature oscillation system that uses constant DC current to produce (B) quadrature signal with four states. (C) An origami crawling robot that leverages the four-state quadrature signal for coordinated contraction of SMA actuators to achieve directional crawling.

*Abstract*— Origami-inspired robots offer rapid, accessible design and manufacture with diverse functionalities. In particular, origami robots without conventional electronics have the unique advantage of functioning in extreme environments such as ones with high radiation or large magnetic fields. However, the absence of sophisticated control systems limits these robots to simple autonomous behaviors. In our previous studies, we developed a printable, electronics-free, and self-sustained oscillator that generates simple complementary square-wave signals. Our study presents a quadrature oscillation system capable of generating four square-wave signals a quarter-cycle out of phase, enabling four distinct states. Such control signals are important in various engineering and robotics applications, such as orchestrating limb movements in bio-inspired robots. We demonstrate the practicality and value of this oscillation system by designing and constructing an origami crawling robot that utilizes the quadrature oscillator to achieve coordinated locomotion. Together, the oscillator and robot illustrate the potential for more complex control and functions in origami robotics, paving the way for more electronics-free, rapid-design origami robots with advanced autonomous behaviors.

## I. Introduction

Origami robots are autonomous machines constructed by folding planar sheets into three-dimensional structures [1]. This top-down design and fabrication strategy offers distinct advantages, most notably rapid and accessible design, prototyping, and manufacturing [1], [2]. In addition, origami robots are typically low-cost [3], lightweight [4], easily storable and deployable [5], and can be scaled across a range of applications [6]. Recent demonstrations have highlighted a broad spectrum of functionalities, including locomotion [7], manipulation [8], shape morphing [9], self-folding [10], computing [11], and multifunctional integration [12], [13].

Despite these advantages, most origami robots rely on bulky electronic components for actuation and control [14], [15]. Integrating compliant origami structures with rigid electronics introduces challenges in design and assembly [16]. More critically, this dependence on electronics restricts the applicability of origami robots in harsh environments, such as regions with high radiation or strong magnetic fields, where semiconductor-based systems might fail [17]. The ability to operate in such regions is especially important for locomotion-based robots designed to expand exploration capabilities and access hazardous terrains.

Research on electronics-free origami robots, however, remains limited due to the challenges of achieving control without electronics [18], [19]. In our earlier work, we developed a self-sustained origami oscillator that generates periodic oscillations from constant electrical power [18]. While this is a major step towards electronics-free control, the oscillator's simple complementary square-wave output constrains the robot to basic autonomous behaviors, such as asymmetric crawling or swimming controlled by one oscillator [4], [18].

In origami robotics, realizing advanced functions requires control strategies that extend beyond single on/off actuations. In particular, quadrature signals—square waves offset by a quarter cycle (Fig. 1A)—are essential for a variety of applications, especially locomotion. For instance, peristaltic motion in worm-inspired origami robots requires multiple phase-shifted waves to sequentially actuate body segments [20]. Myriapod-inspired origami robots similarly employ quadrature signals to coordinate leg-lifting and leg-rotation for effective movement [21]. High-speed quadrupedal origami micro robots also depend on quadrature signals to synchronize the two degrees of freedom in each leg, enabling rapid and coordinated locomotion [22]. The simultaneous need for sophisticated control in origami robots and the advantages of electronics-free designs—particularly their ability to operate in extreme environments—highlights the demand for a new class of electronics-free origami robots capable of producing complex control signals to enable advanced autonomous behaviors.

In this study, we present a printable, self-sustained oscillation system capable of generating quadrature signals from constant electrical power (Fig. 1B). The design builds on our previous single-oscillator system [18] and achieves quadrature generation by coupling three oscillators, where a central oscillator coordinates the output of two peripheral oscillators. The resulting system is configurable, durable, and low-cost, and it can autonomously drive multiple actuators to enable coordinated actuation.

[1] Department of MAE, UCLA. [2] Department of EE, UCLA. [3] Department of MAE, University of California, Davis.
† Corresponding author, wyan@ucdavis.edu

To demonstrate the value and practicality of the oscillation system, we design and construct an electronics-free origami crawling robot that exploits quadrature signals for coordinated locomotion (Fig. 1C). Locomotion is achieved by cycling through the four distinct states of the quadrature signal, which sequentially deform the robot's structure. These asymmetries modulate the normal (and therefore frictional) forces between the legs, resulting in net forward motion.

Together, the oscillation system and robot demonstrate the feasibility of achieving more complex control in origami robotics, advancing the path toward electronics-free, rapidly designed robots with more advanced autonomous behaviors.

The contributions of this paper are: (i) An electronics-free origami oscillation system that generates quadrature signals, verified by qualitative reasoning and experiments. (ii) A crawling origami robot that uses quadrature signals to achieve coordinated motion, verified through theory and experiments.

## II. ORIGAMI SELF-SUSTAINED OSCILLATION SYSTEM

### A. Single Oscillator

The single self-sustained oscillator serves as the fundamental building block of the oscillation system (Fig. 2A). It consists of a bi-stable buckled beam with two stable equilibrium states, two biased contacts, and a pair of shape memory alloy (SMA) fiber actuator. The SMA fiber actuators are connected to biased contacts on one end and to a positive voltage at the center. Ground is applied to the copper plates on both ends for which the biased contacts may touch.

Oscillation is achieved through alternating snap-through transitions of the bi-stable beam. In State 1 (Fig. 2B), the left biased contact touches the grounded copper plate, closing the circuit and activating the left SMA actuator. Contraction of this actuator drives the beam to snap into State 2 (Fig. 2B). In State 2, the right biased contact completes the circuit, activating the right SMA actuator and returning the beam to State 1. This cycle repeats, producing sustained oscillation. A detailed mechanism design and analysis, including repeatability and robustness, can be found in [18].

The previous design employed conductive supercoiled polymer (CSCP) actuators to drive the bi-stable beam transitions. However, the CSCP's slow cooling rates necessitated an external cooling system to achieve meaningful oscillation frequencies, limiting the applicability of the design in real-world environments. In the present design, we replace CSCP actuators with SMA fibers (BMF150) of 0.15 mm diameter, whose rapid heat dissipation eliminates the need for external cooling. This improvement transforms the oscillator from a laboratory demonstration into a self-sufficient component more suitable for deployment in realistic environments.

Experimental characterization further shows that the heating rate of the SMA fiber actuator is predictable and strongly current-dependent, allowing the oscillator to operate over a tunable range of periods (2.2–6.0s). Fig. 3 shows the displacement curves of the bi-stable beam (based on the movement of the center point) for a range of input currents; the data was extracted from the video analysis software Tracker.

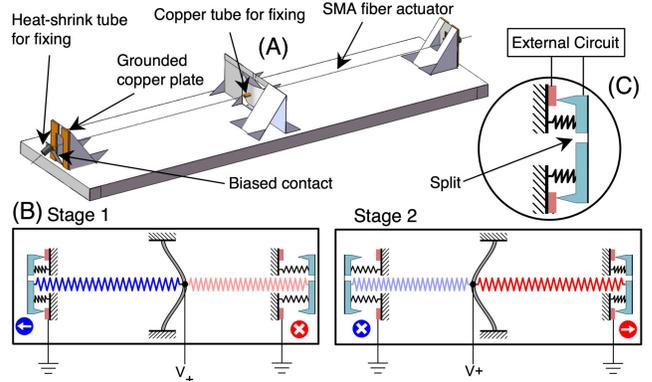

Figure 2. Operation of a single oscillator. (A) Oscillator design. (B) Two stages of the oscillation cycle. (C) Close-up view of biased contact.

Fig. 3 further shows the average oscillation period $T_{avg}$ and its standard deviation $\sigma_T$ for four input currents (0.23-0.26A), measured over 15 consecutive oscillations. The results confirm that the oscillation period decreases with increasing current, enabling "pre-programmable" autonomous robot behaviors through current tuning. In future work, employing SMA fibers of varying diameters (e.g., BMF100 and BMF75) could further expand the oscillator's frequency range.

### B. Oscillation System

The quadrature oscillation system is formed by coupling a central oscillator with two peripheral oscillators. As shown in Fig. 2C, each biased contact on the oscillator is split into two electrically insulated halves, enabling the completion of two independent circuits upon contact with the grounded copper plate. For the central oscillator, half of each biased contact completes its own circuit, while the other half connects to the circuits of the side oscillators (Fig. 4).

When a biased contact for the central oscillator touches the ground, the circuit for one peripheral oscillator is closed, activating its SMA fiber actuator and triggering a snap-through transition of its bi-stable beam. The central oscillator then switches state, completing the circuit for the opposite peripheral oscillator and activating its snap-through. This repeating sequence generates four quadrature signals to control four actuators. Fig. 4 shows the eight distinct stages in each cycle together with graphs that illustrate the fraction of quadrature signal generated in each stage.

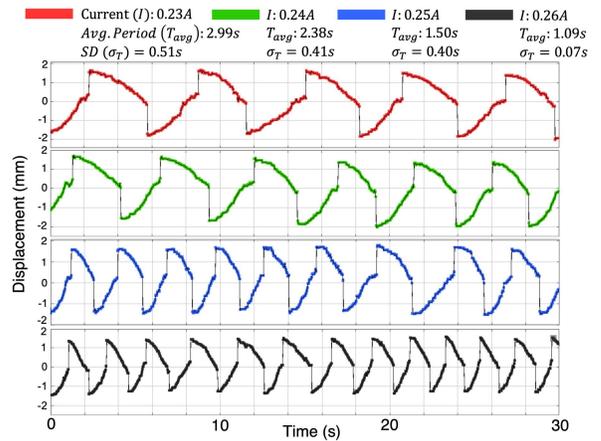

Figure 3. Oscillation displacement curves of the bi-stable beam at different currents (0.23–0.26 A), with average periods and standard deviations.

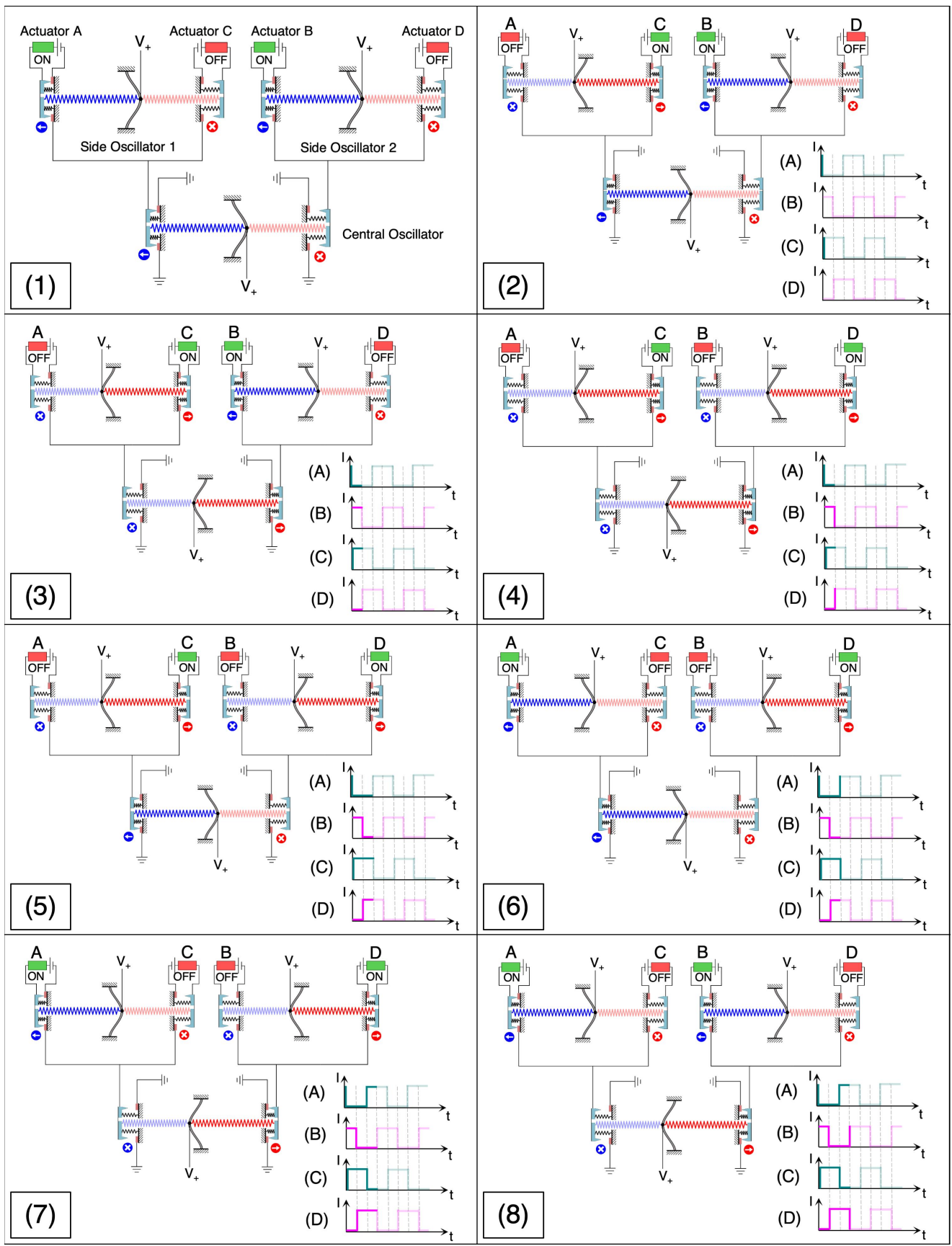

Figure 4. Operation of the oscillation system illustrated over eight stages, with the four quadrature square waves generated at each stage.

A key feature of this coupling scheme is that each oscillator retains its own circuit. Because oscillation frequency is highly dependent on input current, we can independently adjust the period of each oscillator. Ideally, all oscillators operate with identical period with no variation, so that each side oscillator snaps through precisely when the central oscillator switches state, producing perfect quadrature signal. In practice, however, environmental disturbances (e.g., ambient airflow) cause uneven heating and cooling of the SMA fiber, leading to slight period variations. While this problem can be fundamentally addressed by encapsulating SMA fiber actuators with insulating layers to reduce thermal variability, our circuitry scheme offers further buffers. By supplying a higher current to the side oscillators, we allow them to operate with a slightly shorter period than the central oscillator, compensating for its timing variations and ensuring that each side oscillator snaps through exactly once before the central oscillator changes state.

Finally, by connecting external actuators to the two peripheral oscillators, we obtain quadrature actuation—a fundamental coordination pattern widely utilized in robots capable of complex locomotion.

Fig. 5 tracks the displacement curves of the bi-stable beams in each oscillator. Each bi-stable beam controls two actuators by generating two complementary control signals. In total, four quadrature signals are generated. To compute the average phase difference of the quadrature square waves ($\Delta\phi_{avg}$), we calculate the average time offset corresponding to rising and descending edge of the square waves ($\Delta t_{avg}$) and normalize it by the average period $T_{avg}$:

$$\Delta\phi_{avg} = (360° \cdot \Delta t_{avg}) / T_{avg} \quad (1)$$

Where

$$T_{avg} = \Sigma T_i / N \quad (2)$$
$$\Delta t_{avg} = \Sigma(t_{2,i} - t_{1,i}) / N \quad (3)$$

Here, $T_i$ is the period of the i-th cycle, $t_{1,i}$ and $t_{2,i}$ are the corresponding edge times of the two oscillators, and N is the total number of cycles analyzed.

The measured phase offset for an oscillator operating at 0.24A over four cycles is $\Delta\phi=84 \pm 8$. This corresponds to a 6.7% deviation from the ideal 90-degree quadrature relationship; the observed phase variability (standard deviation = 8) corresponds to a relative uncertainty of 8.9%. As explained above, the primary source of error is likely non-uniform heating and cooling of the SMA fiber elements due to environmental disturbances (e.g., ambient airflow). This problem represents engineering refinements rather than fundamental limitations and can be systematically mitigated through thermal insulation strategies. Despite the deviations, these results provide direct experimental validation of quadrature signal generation, establishing the oscillator's ability to produce mechanically derived, phased control signals without electronics.

### III. CRAWLING ROBOT

#### A. Design and Fabrication

The crawling robot consists of a folded PET origami body shaped as an isosceles trapezoid, two BMF150 SMA fiber actuators for forward actuation, and two BMX150 SMA springs for reverse actuation (Fig. 6). The design draws inspiration from robots that achieve locomotion through frictional imbalances induced by geometric asymmetries (e.g., [4], [23], [24]). Forward motion is generated by sequential contraction of the BMF and BMX actuators, which rotate the two legs and introduce geometric asymmetries (Fig. 7). These asymmetries alter the frictional forces between the legs, producing a net forward displacement.

The robot is designed to operate under quadrature actuation, where each leg undergoes coordinated, phase-shifted motion. A key advantage of this design is that locomotion is purely signal driven, enhancing versatility. The more sophisticated quadrature actuation eliminates the need for inherent structural asymmetries such as fixed angled legs for producing directional friction employed in origami robots with simple on/off controls [4]. This represents a step toward more practical locomotion robots, which should not depend on built-in asymmetries that constrain its versatility. For instance, robots relying on directional friction cannot achieve bi-directional motion, whereas our robot can theoretically reverse its direction simply by reversing the control signal sequence.

Further, compared to multi-legged robots that require coordinated leg-lifting and leg-rotation through multiple degrees of freedom and complex mechanisms [21], [22], our design achieves locomotion with a far simpler architecture. This simplicity is particularly advantageous for small-scale robots, where fabrication challenges and payload limitations make multi-DOF designs difficult to implement.

#### B. Simplified Analytical Model

To verify the feasibility of locomotion through geometric asymmetries, we develop a simplified analytical model. In Stage 1 (Fig. 7), the robot is symmetric, and the center of mass lies along the geometric centerline, resulting in equal normal forces on both legs. In Stage 2, as the front leg rotates, the center of mass shifts rearward. Taking moments about the center of mass yields unequal normal forces, with the front leg experiencing a reduced load compared to the back leg. Assuming a uniform floor friction coefficient, this imbalance reduces the frictional force on the front leg, causing the robot to slide forward.

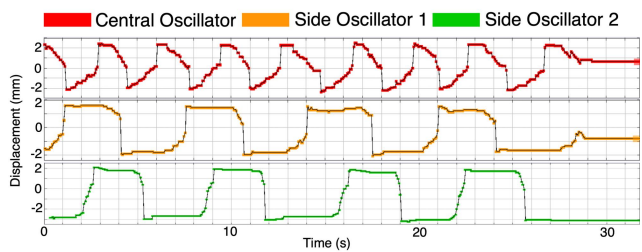

Figure 5. Displacement tracking of the full oscillation system, with the positions of the central and side oscillators plotted together.

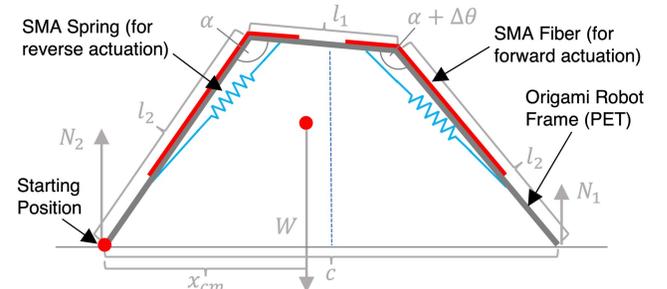

Figure 6. Design and kinetics analysis diagram of the crawling robot.

The normal forces on the front and back legs, $N_1$ and $N_2$, are derived using geometric and statics analysis to be:

$$N_1 = (x_{cm} / c) \cdot W \quad (4)$$

$$N_2 = (1 - (x_{cm} / c)) \cdot W \quad (5)$$

Where c is the distance between the front and back leg, $x_{cm}$ is the distance from the back leg to the center of mass, and W is the total weight. Both c and $x_{cm}$ can be expressed purely as a function of $l_1$, $l_2$, $\alpha$, and $\Delta\theta$. As illustrated in Fig. 6, $l_1$ is the length of the robot's top edge, $l_2$ is the leg length, $\alpha$ is the initial angle between the top edge and the legs, and $\Delta\theta$ is the change in angle during leg extension.

Fig. 7 illustrates the evolution of the normal force distribution over a full locomotion cycle. Each stage produces geometric asymmetries that contributes to forward displacement rather than idle or regressive transitions. For instance, in Stage 1 and 2, $N_1 < N_2$, so the reduced friction on the front leg leads to forward sliding. But in Stage 3 and 4, the relationship reverses ($N_2 < N_1$), so the higher friction on the front leg pulls the robot forward. Thus, every stage of the cycle contributes positively to locomotion, validating the theoretical effectiveness of the proposed mechanism.

Finally, the displacement per cycle d is derived as:

$$d = 2l_2[\sin(\alpha + \Delta\theta - 90°) - \sin(\alpha - 90°)] \quad (6)$$

For our design, this yields a predicted displacement of d=26.6 mm per cycle.

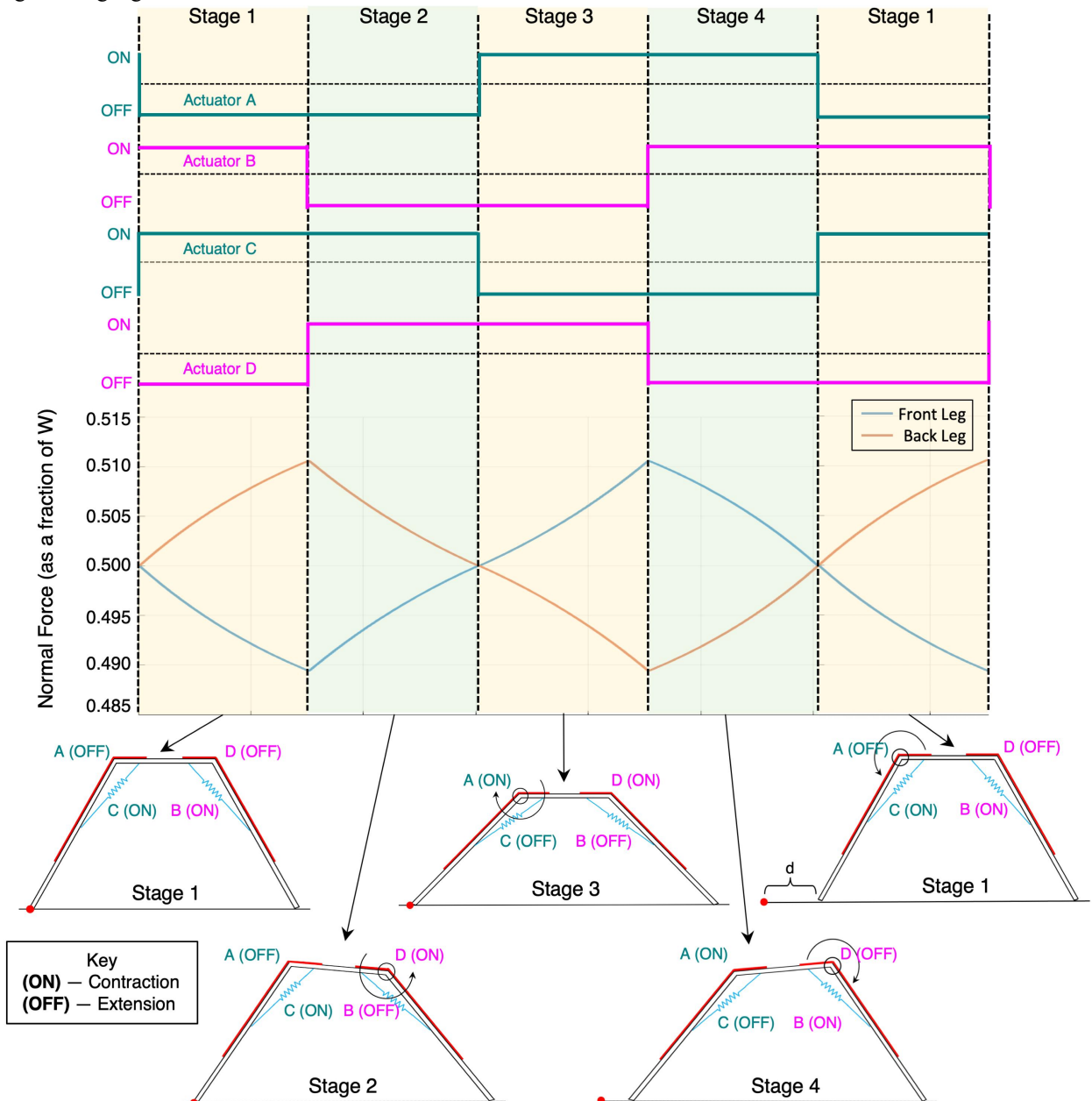

Figure 7. Normal force distribution on the robot's two legs across the four quadrature states, alongside the corresponding robot configuration in each state.

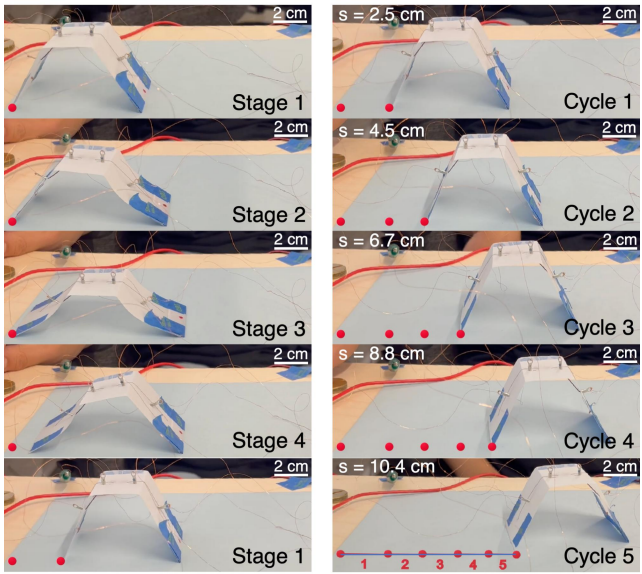

Figure 8. Experimental demonstration of the crawling robot across each stage of motion, manually actuated over five cycles.

*C. Locomotion and Analysis*

Fig. 8 illustrates the geometric structure of the crawling robot and its sequential movements through the five quadrature stages in an experimental demonstration, which closely match the configurations predicted by the theoretical model. The figure also presents five complete locomotion cycles. In this initial demonstration, the actuators were manually triggered in a quadrature sequence to emulate oscillator-driven control, and small pieces of tape were added to introduce minor frictional asymmetries that enhanced locomotion.

The robot achieves an average forward displacement of $d_{avg}$=20.8 ± 0.3mm per cycle, corresponding to a backsliding ratio of 21.8% per cycle. The backsliding results from tension exerted by copper wires and slightly different leg angles from minor folding imprecision. In the future, backsliding may be reduced through optimizing leg-tip geometry to enhance overall frictional contact, thereby rendering folding-induced frictional asymmetries and wire tension negligible. More sophisticated designs can also incorporate structural features that precisely constraints leg angles, and thinner copper wires may be used to further reduce disturbances. Despite these limitations, the demonstration confirms the central result: the robot successfully reproduces the theoretically predicted locomotion patterns required for quadrature-driven coordinated motion.

## IV. CONTROL OF THE CRAWLING ROBOT USING THE OSCILLATION SYSTEM

The final step is to integrate the oscillation system with the crawling robot to demonstrate autonomous, quadrature-driven locomotion. In this setup, the oscillation system and robot are mounted on separate platforms and connected via thin copper wires, with side oscillator 1 driving the front leg and side oscillator 2 driving the back leg. As shown in Fig. 9, the experimental demonstration captures all five stages of the crawling sequence alongside the corresponding oscillator states. A full demonstration showing all stages of the oscillation system is provided in the supplementary video.

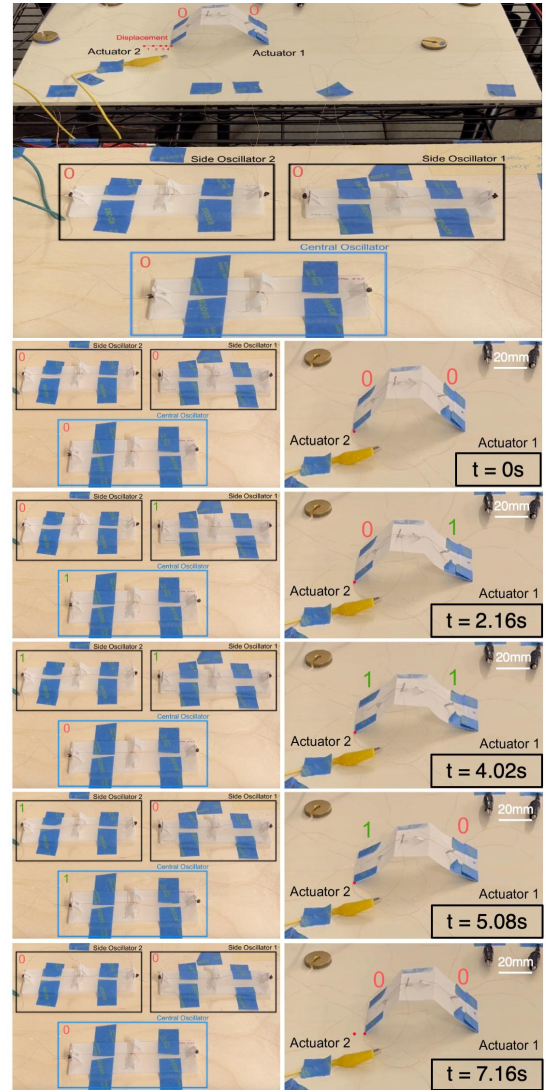

Figure 9. Experimental demonstration of the crawling robot controlled by the oscillation system, showing all locomotion and oscillator states.

Fig. 10 illustrates the quadrature relationship by plotting the displacement curves of the bi-stable beams and crawling robot, with the central oscillator driven at an input current of 0.24 A. The results show that the front and back legs, driven by the two side oscillators, remain consistently about a quarter-cycle out of phase ($\Delta\phi$=84 ± 8° as quantified in Section II)—providing a direct visualization of coordinated quadrature actuation in practice. This alignment further confirms that the oscillator's electrical outputs translate directly into coordinated mechanical motion.

Under control of the oscillation system, the crawling robot achieves an average displacement of 6.9 ± 2.3mm per cycle, and an average speed of 1.3 ± 0.4mm/s, averaged over the front and back leg. While the displacement per cycle is lower than that obtained under manual actuation (20.8mm) and exhibits greater variability, these differences are readily explained by the current system's constraints. Specifically, the average oscillation period is too short for the SMA actuators—particularly the SMA springs—to fully heat and extend, resulting in incomplete leg actuation. In the absence of spring contribution, the robot still functions with the same locomotion pattern, but retraction relies on the folded-memory property of the PET structure rather than complete spring recovery, reducing displacement and repeatability. But crucially, this limitation is not inherent and can be systematically addressed: as shown in Section II, the oscillation period is tunable through the input current. Using thicker SMA fiber actuators with lower heat dissipation would extend the tunable range, allowing sufficient heating time for full spring actuation and improved displacement per cycle. An alternative solution is using thinner SMA spring with shorter actuation time. Since manual activation produced a large displacement of 20.8mm when given adequate heating time and complete leg actuation, these results show that with extended oscillation period, the displacement can be significantly improved.

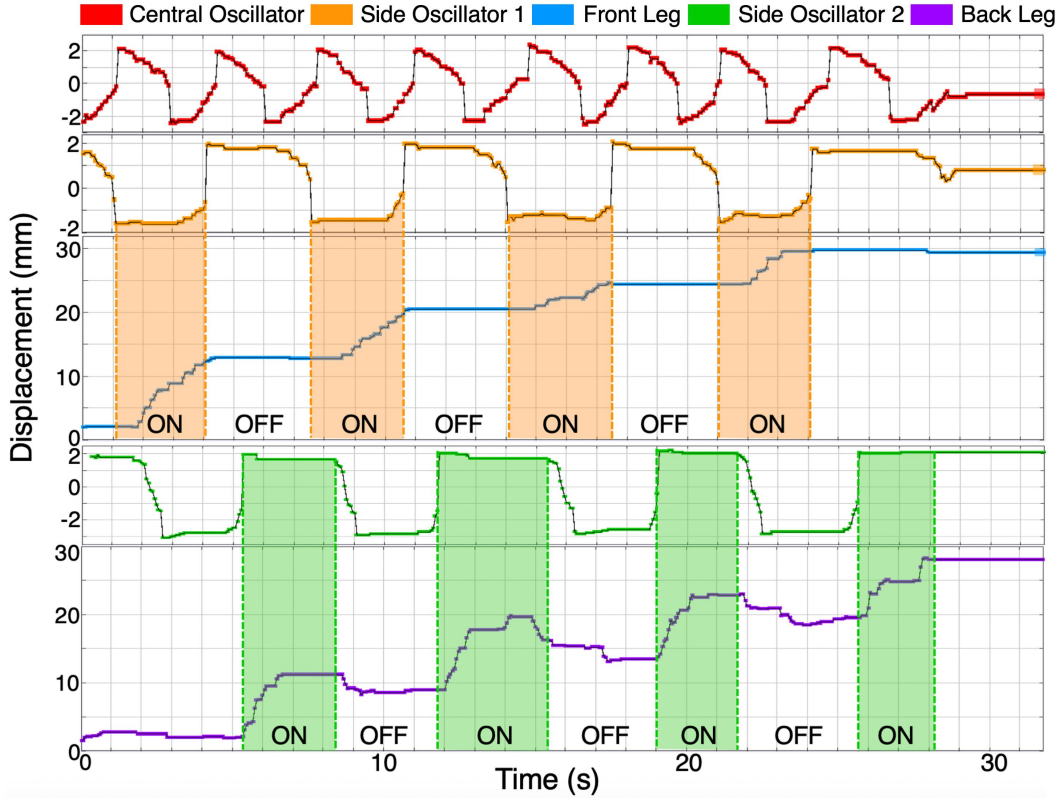

Figure 10. Displacement tracking of the oscillation system controlling the crawling robot over four locomotion cycles. "ON" state corresponds to leg contraction while "OFF" state corresponds to leg extension.

The increased variability in displacement primarily arises from phase deviations in the oscillation system, likely caused by environmental influences such as uneven heating and cooling from ambient airflow. As described in Section III, these effects can be systematically addressed through encapsulation strategies.

Although the present system has limitations—including backsliding during locomotion, the need for frequency tuning to achieve optimal performance, and opportunities to improve phase robustness—the central contribution of this work lies in demonstrating, for the first time, that mechanically generated quadrature signals can directly drive coordinated locomotion in an electronics-free origami robot. This proof-of-concept establishes the feasibility of more complex oscillator-driven locomotion without reliance on bulky electronics such as microcontrollers or motors.

## V. Future Research

### A. Technical Directions

Future work will focus on enhancing practical performance. Priorities include mitigating backsliding and disturbances during locomotion, extending the oscillation period to maximize actuator actuation and crawling speed, and improving system robustness to maintain consistent quadrature signal generation. These refinements will improve locomotion efficiency and oscillation reliability, creating a stronger platform for more advanced demonstrations.

### B. Long-term Vision

The oscillation framework provides a foundation for richer forms of mechanical control. Expanding the system with additional oscillators could enable the generation of smaller phase-shifted signals, such as up to eight square waves separated by one-eighth of a cycle (Fig. 11A), thereby allowing coordination of a larger number of actuators and increasingly sophisticated functions. In parallel, our ongoing development of more compact oscillators—one already 36% shorter than the current version (Fig. 11B)—points toward embedding control directly within the robot body. This trajectory ultimately advances toward fully integrated, electronics-free origami robots with on-board control.

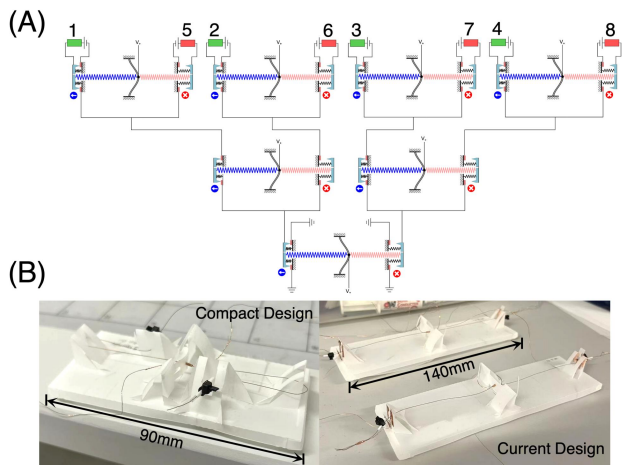

Figure 11. (A) Oscillation system schematic for generating square-wave signals offset by one-eighth of a cycle. (B) Compact oscillator design that is 36% shorter than the current version.

## VI. Conclusion

This work demonstrates the first quadrature oscillation system in electronics-free origami robots. By coordinating three single oscillators, the system enables more complex, phased signal without the use of electronics. Through integration with a crawling robot, we established that the electrical outputs of the oscillation system directly translate into coordinated mechanical actuation, demonstrating more advanced oscillator-driven locomotion.

Although performance optimization—such as reducing backsliding, tuning driving periods, and enhancing robustness—remains ongoing, our work shows that these limitations are both qualitatively understood and practically addressable, underscoring the feasibility of further refinement. More importantly, the proof-of-concept is clearly established: oscillator-driven quadrature control offers a promising, experimentally validated strategy that broadens the design space for electronics-free origami robotics.

Looking ahead, the oscillator framework points toward mechanically intelligent robots capable of far richer behaviors. Scaling the system with additional oscillators could enable multi-phase signal generation for more sophisticated behaviors, while ongoing efforts in miniaturization open the path toward embedding control directly within a robot's body. Together, these directions advance toward fully integrated, electronics-free origami robots capable of autonomous, coordinated motion driven entirely by mechanical computation.

## Acknowledgment

W.Y. thanks the startup funds provided by the Department of Mechanical and Aerospace Engineering, University of California, Davis.